\documentclass[runningheads]{llncs}
\usepackage{graphicx}
\usepackage{amsmath,amssymb} 
\usepackage{color}
\usepackage[T1]{fontenc}
\usepackage[utf8]{inputenc}
\usepackage{authblk}
\usepackage{psfrag}

\begin{document}
\pagestyle{headings}
\mainmatter
	
\def\ACCV18SubNumber{***}  

\title{Prostate Segmentation using 2D Bridged U-net} 
\authorrunning{Prostate Segmentation using 2D Bridged U-net}
\author{Wanli Chen$^{1,+}$, Yue Zhang$^{1,3,+}$, Junjun He$^2$, Yu Qiao$^{2,*}$, Yifan Chen$^{1,4,*}$,Hongjian Shi$^{1,*}$, Xiaoying Tang$^{1,*}$}
\institute{$^1$ Southern University of Science and Technology\\$^2$  Shenzhen Institutes of Advanced Technology, Chinese Academy of Sciences
\\$^3$ The University of Hong Kong
\\$^4$ The University of Waikato
\\ $^+$ Equal contribution
\\ $^*$ Corresponding authors (yu.qiao@siat.ac.cn, yifan.chen@waikato.ac.nz, \\ shihj@sustc.edu.cn, tangxy@sustc.edu.cn) }

%

\maketitle

\begin{abstract}
In this paper, we focus on three problems in deep learning based medical image segmentation. Firstly, U-net, as a popular model for medical image segmentation, is difficult to train when convolutional layers increase even though a deeper network usually has a better generalization ability because of more learnable parameters. Secondly, the exponential ReLU (ELU), as an alternative of ReLU, is not much different from ReLU when the network of interest gets deep. Thirdly, the Dice loss, as one of the pervasive loss functions for medical image segmentation, is not effective when the prediction is close to ground truth and will cause oscillation during training.
To address the aforementioned three problems, we propose and validate a deeper network that can fit medical image datasets that are usually small in the sample size. Meanwhile, we propose a new loss function to accelerate the learning process and a combination of different activation functions to improve the network performance. Our experimental results suggest that our network is comparable or superior to state-of-the-art methods.
\end{abstract}

\section{Introduction}
\subsection{Network for Image Segmentation}
Convolutional neural network shows a great advantage over traditional methods in computer vision. Recently, fully convolutional network (FCN) \cite{FCN} has become the main framework for image segmentation task. Specifically, for medical image segmentation, a popular FCN is U-net \cite{ronneberger2015Unet}. U-net has an encoder-decoder structure with concatenation being used to merge features.  It is widely used in medical image segmentation \cite{vnet} \cite{3dUnet} because of its efficiency and the adaptivity for small dataset. The main drawback of U-net is that it is difficult to go very deep. To fully exploit the utility of U-net but go deeper, a stacked U-net has been proposed. However, such network is likely to be trapped into a sub-optimal solution because a stacked U-net is more complicated than a single U-net. As such, a stacked U-net is usually employed when there is a pre-train model \cite{shah2018stacked} \cite{ghosh2018stacked}or fed with a large number training data (more than 10K) \cite{ma2018docunet}. There is a work concatenating two U-net by two loss \cite{xia2017w}, but they didn't consider the information sharing in two U-net.In this paper, we propose a bridging architecture between two U-nets. Specially, we connect each decoder layer of the first U-net with the corresponding encoder layer of the second U-net, which directly inputs the features of the previous layers into the latter layers. This process reduces the training cost and exhibits a better performance than a single U-net. By using network bridging, a stacked U-net can deal with small datasets and be used in medical image segmentation without a pre-train model.
\subsection{Feature Fusion Methods}
To bridge two U-nets, an appropriate method is needed. Network bridging can also be viewed as feature fusion. The main fusion methods can be divided into two categories: addition and concatenation. Addition is an intuitive way. It directly adds features together. This fusion method has been widely used in many computer vision tasks such as ResNet \cite{resnet} based classification,  Feature Pyramid Network (FPN) \cite{FPN} based detection and FCN based image segmentation. This method can be viewed as highway gradient transfer \cite{highwayNET}, which will accelerate gradient propagation. Addition will change the distribution of weights, which is pernicious for network initialization. This issue can be alleviated when using concatenation for feature fusion. Representative networks include DenseNet \cite{densenet} and U-net. We will have a further discussion on concatenation and addition in part 3 of this paper and show that concatenation is superior for our network bridging.
\subsection{Activation Functions}
To make our network performs better, different activation functions are applied in our network. Activation function is an important component of a neural network. The activation layers adds non-linearity to the weights so that the network can deal with more complex tasks. Previously, sigmoid \cite{sigmoid} has been used as the activation function. However, sigmoid will saturate during training. Then ReLU \cite{ReLU} has been proposed to solve the saturation issue. When using ReLU as the activation function, the learning rate should be carefully adjusted because ReLU gets saturated in negative axis, and a big learning rate will “kill” some neurons. To address this efficiency, Exponential ReLU (ELU) \cite{ELU} has been proposed. ELU does not get saturated in the negative axis immediately. However, saturation still happens when the network gets deeper. When ELU gets saturated, it is no different from ReLU. As such, we can replace some ELU layers with ReLU. Because ELU is not saturated when the negative axis is 0, the replacement could be viewed as “reset” ELU, which will re-activate the subsequent saturated ELU neuron. In this paper, we use ELU and ReLU simultaneously. We also designed a new loss function for medical image segmentation.

To sum up, we have three contributions. Firstly, we propose a new network structure to accelerate the learning process of stacked U-net. Additionally, we investigate the performance of different feature fusion methods. Secondly, we explore the utility of using ELU and ReLU as the activation functions to reach a superior performance. We also extend the result into the general image classification task. Thirdly, we design a new loss function.

\section{Related Work}
\subsection{U-net}
The U-net consists of a down-convolutional part and up-convolutional part. The down-convolutional part aims at extracting features for classifying each voxel into one or zero. It consists of repeated application of two $3\times3$ convolutions. At each downsampling step the number of feature channels is doubled. And the up-convolutional part aims at locating regions of interest (ROI) more precisely. Every step in up-convolutional part consists of an upsampling of the feature map followed by a $2\times2$ convolution that halves the number of feature channels, a concatenation with the correspondingly cropped feature map from the contracting path, and two $3\times3$ convolutions. Max pooling and ReLU activation was used for the convolution block in U-net.

\subsection{PROMISE12}
Determination of prostate volume (PV) can help detect pathologic stage of diseases, such prostate cancer. What's more, the accurate prostate specific antigen (PSA) is dependent on the quality of the PV. The accuracy and variability of PV determinations pose limitations to its usefulness in clinical practice. It is also an essential part in clinical to get the the size, shape, and location of the prostate relative to adjacent organs. Recently, this kind of information can be obtained by MRI using the high spatial resolution and soft-tissue contrast. This, combined with the potential of MRI to localize and grade prostate cancer, has led to a rapid increase in its adoption and increasing research interest in its use for this application. Consequently, there is a real clinical and research need for the accurate robust, automatic prostate segmentation methods used as an pre-processing procedure for computer-aided detection and diagnostic algorithms, as well as a number of multi-modality image registration algorithms.

\par
\begin{figure}[htbp]
    \centering
    \includegraphics[width=8cm]{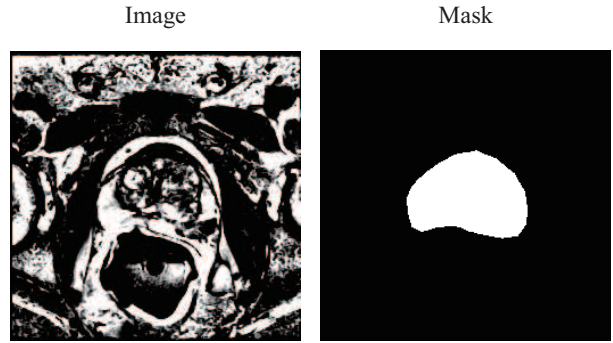}
    \caption{A sample in PROMISE12. Mask means the ground truth segmentation result, which is only available in training data.}
    \label{fig:PROMISE}
\end{figure}

 Prostate MR Image Segmentation challenge 2012 (PRPMISE12) was held to compare segmentation algorithms for MRI of the prostate during MICCAI 2012. After MICCAI 2012, the organizer are still receiving and uploading submission \cite{litjens2014evaluation}. For training data, the MRI images and ground truth segmentation results are all available, shown in Fig. \ref{fig:PROMISE}. However, the ground truth segmentation results are only available for organizers, who will evaluate the results of participants. In this paper, we use PROMISE12 dataset as our training and validation dataset and we also submitted our testing results to organizers.


\section{Model Setup}

\subsection{Bridged U-net}
\begin{figure}[htbp]
    \centering

    \includegraphics[width=12cm]{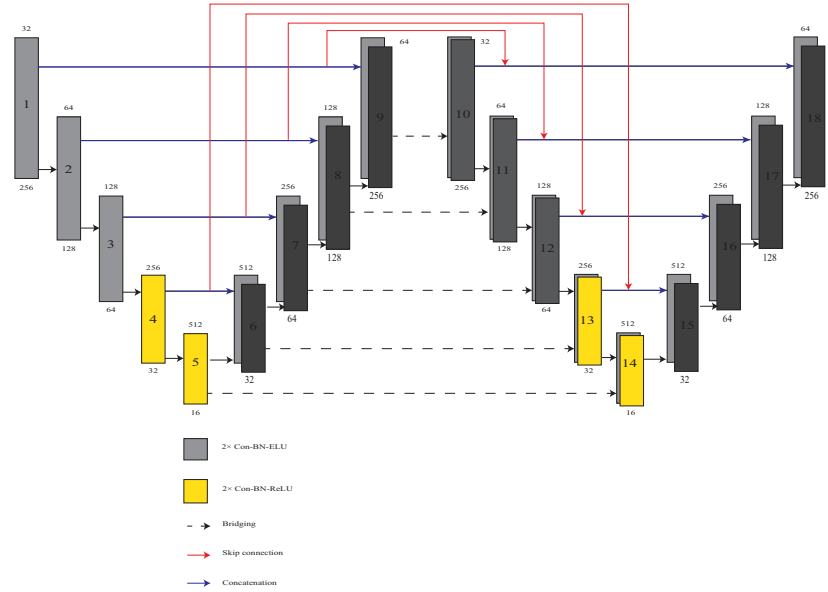}
    \caption{Bridged U-net architecture. The number above each block represents the number of feature channels. The number inside each block represents the sequence number. The number below each block means the image size.}
    \label{fig:model}
\end{figure}

Our network is based on U-net, which is a classical encoder-decoder net in medical image application. Based on U-net, a stacked U-net is proposed. The stacked U-net improves network performance by using the first U-net to find a coarse feature and use the second U-net to obtain a fine result. The stacked U-net is, however, not useful for medical image segmentation. It is hard to reach convergence and usually dive into a sub-optimal solution because the increasing complexity of network. To overcome the issue, we propose a network bridging method. Different from the previous stacked U-net which acquires large number training data, bridging two U-nets can reduce the training cost and makes the network fit for medical application where the training data are usually not sufficient. This is because bridging two U-nets can fully use different features in multi levels, which will accelerate the convergence of neural network. Our network structure is shown on Fig. \ref{fig:model}. The gray block represents a ELU cluster (2 conv-BN-ELU blocks), and the yellow block represents a ReLU cluster (2 conv-BN-ReLU blocks). The dotted lines represents network bridging. The red lines represents skip connections.

\subsubsection{Network Bridging}
In order to bridge features in two U-nets, we can use addition or concatenation as our bridging method. Both methods are widely used in computer vision. For example, FPN used addition to combine low level features with higher level semantic features, while DenseNet uses concatenation for features combination. To merge features of different levels, we argue that concatenation is more effective. We proved that in the perspective of weights initialization.
\par
\begin{figure}[t!]
    \centering
    \includegraphics[width=8cm]{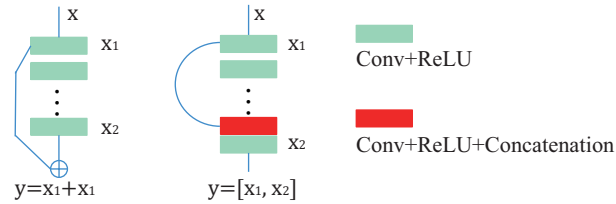}
    \caption{The comparison of addition and concatenation}
    \label{fig:bridging}
\end{figure}

 In weights initialization, to make sure information flow, we need to keep the variance of input equal to output, $Var\left[ x \right]=Var\left[ y \right]$, as shown on Fig. \ref{fig:bridging}.
Assuming all convolution layers are initialized with gaussian normal filler and using ReLU as activation function and all parameters are independently and identically distributed (IID). According to the fact that addition of two normal distribution is another normal distribution with $\sigma=\sigma_1+\sigma_2$ and concatenation of two IID normal distribution is another distribution with  $\sigma=\sigma_1=\sigma_2$. Then we obtain:
\par Addition:
\begin{equation}
  Var\left[y \right]=Var\left[x_1 +x_2 \right],
  \label{add1}
\end{equation}
\begin{equation}
  Var\left[y \right]=2Var\left[x \right],
  \label{add2}
\end{equation}

\par Concatenation:
\begin{equation}
  Var\left[y \right]=Var\left[x_1 , x_2 \right],
  \label{concatenation1}
\end{equation}
\begin{equation}
  Var\left[y \right]=Var\left[x \right],
  \label{concatenation2}
\end{equation}

\setlength{\tabcolsep}{2pt}
\begin{table}
\begin{center}
\caption{Influence of network bridging and skip connection. vDSC is the abbreviation of volumetric Dice Similarity Coefficient. }
\label{table:bridge}
\begin{tabular}{l|c|c|c}
\hline\noalign{\smallskip}
Method & Bridging method & Skip connection & Mean vDSC $\left[ \% \right]$\\
\noalign{\smallskip}
\hline
\noalign{\smallskip}
U-net &None &None &86.73 \\
Stacked U &None &None &85.57 \\
Stacked U & Addition & None &86.99 \\
Stacked U & Concatenation & None &87.85 \\
Stacked U & Concatenation & Concatenation &86.02 \\
Bridged U-net & Concatenation & Addition &\textbf{88.12} \\

\hline
\end{tabular}
\end{center}

\end{table}
\setlength{\tabcolsep}{1.4pt}

\par Therefore, using concatenation can guarantee the information flow. Thus, we deem concatenation is better for feature fusion. This result is also proved in our network by ablation experiments shown on Table \ref{table:bridge}. In this table, we only use ELU as our activation function. The results shows that using concatenation and skip connection can improve network performance.
\subsubsection{Skip connection}
Skip connection is an important part for medical image segmentation, which is helpful to improve network performance. In Bridged U-net, we use addition for skip connection. Although concatenation is more effective in the perspective of weight initialization, addition has it own advantages. As it shown in Fig. \ref{fig:model}, we connect the two U-nets in their concatenation stage. The reason why not using concatenation is that it will cause redundancy. If using concatenation as our skip connection method, the decoder part of the second U-net have to learn more parameters than the first U-net, which aggravate the learning burden of the second U-net and the network will not converge.

\subsection{Activation Function: The Combination of ReLU and ELU}
In artificial neural networks, the activation function play an important role. Rectifier liner unit (ReLU) is the most popular activation function for deep neural network \cite{lecun2015deep}. Exponential  liner unit (ELU) replace the negative part in ReLU with exponential function, which is helpful to make the average of output close to zero \cite{ELU}. In our network, we initially use ELU with all layers.
\par Both ReLU and ELU are widely used in segmentation task. In this work, we find the combination of ReLU and ELU can improve the segmentation performance. Neural networks usually suffer low coverage rate because of vanishing gradient, especially for deep network. ELU provides a buffer in negative axis so that it will not saturate immediately. However, ELU still suffers the saturation problem when network gets deeper.
\par When ELU saturated to negative values, there will be no difference between ELU and ReLU. Therefore, although ELU is used, the performance of our network is close to the network that only use ReLU.
\par To overcome this issue, we have proposed a method using ELU and ReLU simultaneously. Because ELU will be saturated when the network going deeper and has the same effect as ReLU, we can simply replace some ELU layers with ReLU. This replacement will not influence the function of these specific layers since ELU is the same as ReLU on that condition. However, this replacement will affect the following layers. The saturated negative values in the following ELU layers will be reset to 0 because of ReLU, which means these following layers are not saturated anymore. We replaced the activation function of $7^{th}$ - $10^{th}$ and $25^{th}$ - $28^{th}$ convolution layers with ReLU (cluster 4, 5, 13, 14 shown on Fig. \ref{fig:model}). The results in shown on Table \ref{table:np}.
\begin{table}[ht]
\begin{center}
\caption{Network performance using different activation functions. ELU/ReLU only means using ELU/ReLU in all layers. Cluster 1 means replacing the activation function in block 3, 7, 12, 16 shown on Fig. \ref{fig:model} with ReLU. Cluster 2 means replacing the activation function in block 5, 9, 10, 14 shown on Fig. \ref{fig:model} with ReLU. Cluster 3 means replacing the activation function in block 4, 5, 13, 14 shown on Fig. \ref{fig:model} with ReLU. The result shows that it is unwise to add ReLU layer frequently (Cluster 1). Two ReLU blocks should have a relative large interval (Cluster 3).}
\label{table:np}
\begin{tabular}{c|c|c|c|c|c}
\hline\noalign{\smallskip}
ELU only &  ReLU only & Cluster 1 & Cluster 2 &Cluster 3 & Mean vDSC $\left[ \% \right]$\\
\noalign{\smallskip}
\hline
\noalign{\smallskip}
Yes & & &&& 88.12 \\
&Yes & &&& 88.07 \\
& &Yes &&& 87.56 \\
& & &Yes&& 88.10 \\
& & &&Yes& \textbf{89.10} \\
\hline
\end{tabular}
\end{center}

\end{table}
\setlength{\tabcolsep}{1.4pt}

\par However, it is unwise to add ReLU layers frequently. Two ReLU blocks should have a relative large interval. This is because the ReLU layer should only be added when ELU is going saturated. If adding ReLU frequently, it is hard to guarantee the saturation of ELU. In our network, we gap the two ReLU clusters with 18 convolution layers (9 ELU blocks), which is shown on Fig. \ref{fig:model}. Since image segmentation is a pixel-wise image classification task. We want to find whether this method is useful in traditional image classification task.
\par
\begin{figure}[t!]
    \centering
    \includegraphics[width=8cm]{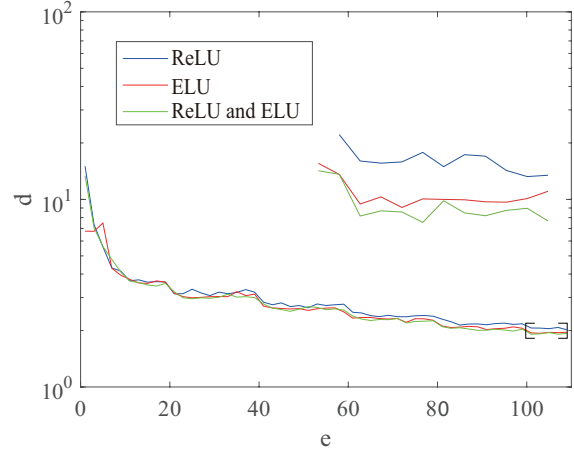}
    \caption{Validation loss of VGG-16 in CIFAR-100 using different activation functions.}
    \label{fig:VGGloss}
\end{figure}

 Then we use CIFAR-100 \cite{cifar100} as our training-validation dataset and VGG-16 \cite{VGG} as our model. In the training phase, the learning rate is initially set as 0.1 and decay 2 times after 20 epochs. We choose 128 as batch size and $10^{-6}$ as weight decay. After 250 epochs training, the result is shown on Table \ref{table:relu}. In this experiment, we also use ELU for all layers but replace the activation function of $4^{th}$ and $5^{th}$ convolution layers with ReLU. The results shows that the replacement can improve the network performance. The validation loss curve is shown on Fig. \ref{fig:VGGloss}, from which we can find that the ELU shows better performance than ReLU initially. Additionally, the performance of ELU and ReLU combination is better than ReLU but worse than ELU on the beginning. The reason is that ELU is not saturated at the beginning. However, with the growth of epochs, the ELU-ReLU combination starts chasing and shows the lowest loss among them. The reason is that ELU starts saturating with the growth of epochs, but ReLU can reset the saturated ELU to 0 so that the following ELU layers are not saturated anymore.

\begin{table}[ht]
\begin{center}
\caption{Accuracy performance of VGG-16 using different activation functions on CIFAR-100 dataset.}
\label{table:relu}
\begin{tabular}{l|c|c|c|c}
\hline\noalign{\smallskip}
Model & ReLU only &  ELU only & ELU and ReLU & Accuracy\\
\noalign{\smallskip}
\hline
\noalign{\smallskip}
VGG-16  &Yes &  & & 0.7052 \\
        & &Yes  & & 0.7163 \\
        & &  &Yes & \textbf{0.7201} \\
\hline
\end{tabular}
\end{center}

\end{table}
\setlength{\tabcolsep}{1.4pt}

\subsection{Cos-Dice Loss Function}
Dice loss, as the most popular loss function for medical image segmentation, uses dice similarity coefficient (DSC) to generate training loss. DSC is a statistic used for comparing the similarity of two sets. It is calculated as this:
    \begin{equation}
    DSC(GS,SEG)=\frac{2 \left| GS \cap SEG \right|}{\left| GS \right| + \left| SEG \right|},
    \end{equation}
where $GS$ represents the gold standard segmentation of a prostate region, $SEG$ represents the corresponding automatic segmentation,and $\left| GS \cap SEG \right|$ refers to the overlap region. $\left| \cdot \right|$ represents the sum of the entries of matrix. The dice loss is defined as
    \begin{equation}
    L_{Dice} = 1-DSC ,
    \end{equation}
\par In medical image segmentation, dice loss is more effective than other loss functions that used in semantic segmentation. Because the number of positives and negatives are highly unbalanced in the task of medical image segmentation. However, the dice loss has its own limitation.
\par
\begin{figure}[t!]
    \centering
    \includegraphics[width=8cm]{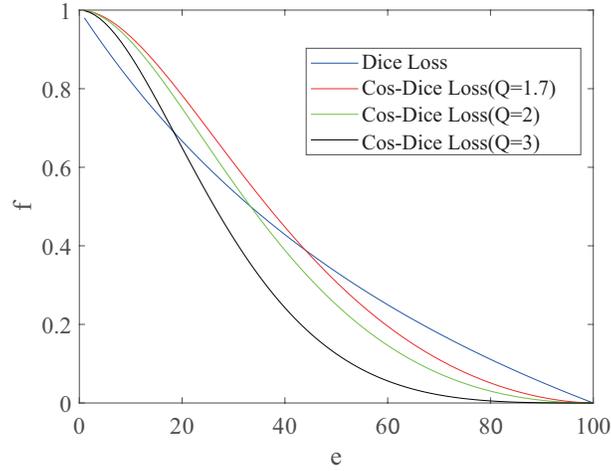}
    \caption{Loss-Intersection graph of dice loss and cos-dice loss with different factor Q}
    \label{fig:cosloss}
\end{figure}

 To illustrate that, we have investigated back propagation function. Assume $z_j^l$ is the $j^{th}$ input of $l^{th}$ Layer, $a_j^l$ is the $j^{th}$ output of $l^{th}$ layer, $\sigma$ is the activation function. Then we obtain
    \begin{equation}
    a_j^l=\sigma \left( z_j^l \right) ,
    \end{equation}

    \begin{equation}
    z_j^l= \sum_{k}w_{jk}^{l}a_k^{l-1} + b_j^l ,
    \end{equation}
We now focus on the last layer. Suppose $\delta_j^l$ is the error and $L$ is the loss function, then we obtain
    \begin{equation}
    \delta_j^l = \frac{\partial L}{\partial z_j^l}=\frac{\partial L}{\partial a_j^l} \cdot \frac{\partial a_j^l}{\partial z_j^l} ,
      \label{errorloss}
    \end{equation}
Thus,
    \begin{equation}
    \delta_j^l \propto \frac{\partial L}{\partial a_j^l}
    \end{equation}
    It can be observed that the error is proportional to the partial derivative of loss function to output. Then we can plot the Loss-Intersection graph for dice loss shown in Fig. \ref{fig:cosloss} using blue line. The intersection percent can be regard as the ratio of  $a_j^l$ in last layer to ground truth.

According to the equation $\left( \ref{errorloss} \right)$, the back propagated error could be calculated as the gradient of Loss-Intersection curve. According to Fig. \ref{fig:cosloss}, we find that gradient of dice loss is not varied. In other words, when error back propagated, there is no much difference between the intersection percentage of 20\% and the intersection percentage of 70\% considering gradient. This deficiency will cause oscillation when the learning rate decrease. To solve the issue, we need to design a loss function that has larger penalty when the intersection area is small and smaller penalty when the intersection area is big.
\par We propose Cos-Dice Loss Function:
    \begin{equation}
    L_{CosDice}=\cos^Q \left(\frac{\pi}{2} \cdot DSC \right), Q>1 .
    \end{equation}
Where Q is an adjustable number. As it shown in Fig. \ref{fig:cosloss}, the cos-dice loss is smoother than dice loss when the intersection percentage is large and rougher than dice loss when the intersection percentage is small.

\begin{figure}[t!]
    \centering

    \includegraphics[width=8cm]{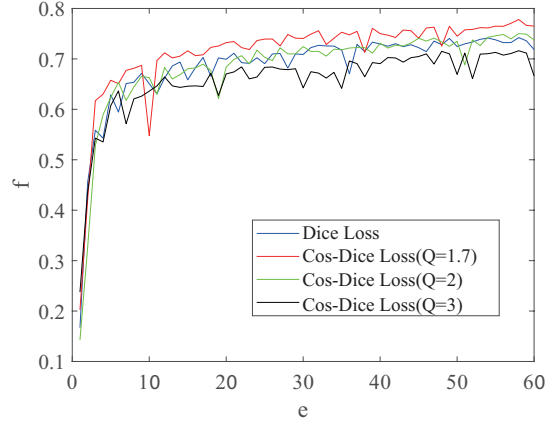}
    \caption{The performance of dice loss and cos-dice loss with different factor Q}
    \label{fig:coslossper}
\end{figure}
\begin{table}[b]
\begin{center}
\caption{Performance of cos-dice loss with different Q.}
\label{table:cos-dice}
\begin{tabular}{l|c}
\hline\noalign{\smallskip}
Loss & Mean vDSC $\left[ \% \right]$ \\
\noalign{\smallskip}
\hline
\noalign{\smallskip}
Dice Loss& 89.10\\
Cos-Dice Loss $Q=1.7$& \textbf{89.56}\\
Cos-Dice Loss $Q=2$& 88.77\\
Cos-Dice Loss $Q=3$& 87.79\\
\hline
\end{tabular}
\end{center}

\end{table}
\setlength{\tabcolsep}{1.4pt}

By adding cos-dice loss, we have a more stable result with better performance. Table \ref{table:cos-dice} shows that we have 0.46\% gain in performance and the model becomes more stable.
Actually, the principle of cosine transform is adding a weight into original dice loss:
    \begin{equation}
    \begin{split}
    \frac{\partial L_{CosDice}}{\partial a_j^{j}} = -Q \cos^{Q-1} \left( \frac{\pi}{2} DSC\right) \cdot \sin^{Q-1} \left( \frac{\pi}{2} DSC\right) \cdot \frac{\pi}{2} DSC^{'}
    = w \cdot \frac{\partial L_{Dice}}{\partial a_j^{j}},
    \end{split}
    \end{equation}
where $ w=-Q \cos^{Q-1} \left( \frac{\pi}{2} DSC\right) \cdot \sin^{Q-1} \left( \frac{\pi}{2} DSC\right) \cdot \frac{\pi}{2}$. From equation (12) we can see that the back propagated error calculated by cos-dice loss is similar to original dice loss, which makes cos-dice loss maintain the advantages of dice loss. Additionally, this weight is easy to modify by adjusting $Q$. A bigger $Q$ leads to a smoother loss. However, the performance of network will decrease when $Q$ is too big, as it shown on Table \ref{table:cos-dice}. Therefore, $Q$ should be carefully adjusted to obtain the optimal result.

\subsection{Implementation Details}
\subsubsection{Pre-processing}
PROMISE12 challenge provides 50 training datasets, each dataset contains one 3D prostate MRI image that composed of several 2D slices. We choose 45 datasets for training and 5 datasets for validation. The validation dataset number is 5, 15, 25, 35, 45. We simply resize every slice to $256\times256$ as our pre-processing method. The data augmentation was applied by random flipping, rotation from -10$^\circ$ to 10$^\circ$ to generate more data. The original training set contains 1250 slices. We obtained 5000 images (still a relatively small number for stacked U-net) after data augmentation.
\subsubsection{Implementation}
The proposed method was implemented in Python language, using Keras with Tensorflow backend. All experiments are conducted on a Linux machine running Ubuntu 16.04 with 32 GB RAM memory. Bridged U-net is trained using two GTX 1080 Ti GPUs. We use Adam optimizer with initial learning rate 0.001 and 24 batch size for training.
\section{Results}
\subsection{Validation Results}
\subsubsection{Qualitative Comparison}
\begin{figure}[ht!]
    \centering
    \includegraphics[width=8cm]{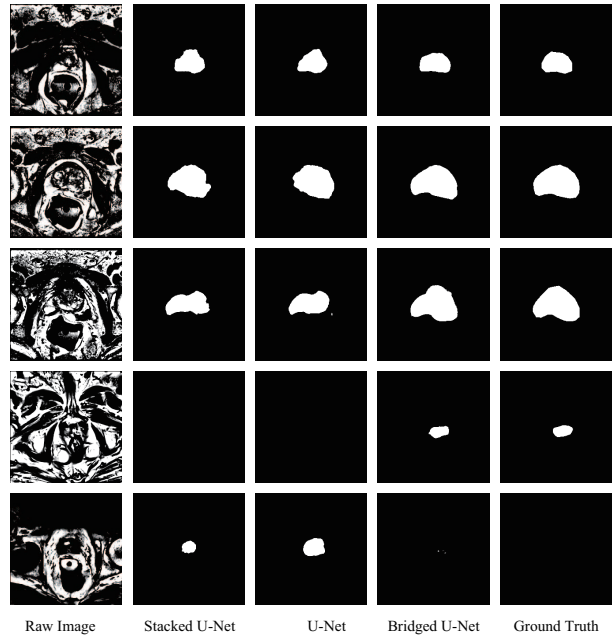}
    \caption{Segmentation results. From left to right are raw image, the segmentation results of U-net, the segmentation results of a stacked U-net, the segmentation results of Bridged U-net, ground truth respectively}
    \label{fig:valimage}
\end{figure}
To intuitively compare the proposed method with the U-net and stacked U-net, the segmentation results of some representative and challenging samples are shown in Fig.\ref{fig:valimage}. It can be observed that these prostate images have fuzzy boundaries and the pixel intensity distributions are inhomogeneous both inside and outside of the prostate.
Additionally, both prostate and non-prostate regions have similar contrast and intensity distributions. All of these phenomenons make the segmentation task difficult.
\par As shown in the second column and third column in Fig.\ref{fig:valimage}, both stacked-U-net and U-net failed to obtain satisfactory result, though the model could detect part
of prostate. The results of Bridged U-net are shown in the fourth column of Fig. \ref{fig:valimage}. The fuzzy boundaries are well detected by our proposed method, Bridged U-net. Besides, the segmentation boundary are more continuous and smooth than the competing method. It can also be observed that the Bridged U-net can reduce false negative and false positive rate according to the fourth and fifth rows in Fig. \ref{fig:valimage}.

\subsubsection{Quantitative Comparison}
\begin{table}[ht]
\begin{center}
\caption{Quantitative comparison between the proposed method with other methods. Abbreviation: (a) vDSC: volumetric Dice Similarity Coefficient, (b) STD: Standard deviation, (c) HD: Hausdorff Distance, (d) ABD: Average Boundary Distance, (e) RAVD: Relative Absolute Volume Difference. $\uparrow$ means the higher value is better. $\downarrow$ means the lower value is better.}
\label{table:valdsc}
\begin{tabular}{llcl}
\hline\noalign{\smallskip}
parameter & Stacked U &  U-net & Wnet\\
\noalign{\smallskip}
\hline
\noalign{\smallskip}
Mean vDSC$\left[ \% \right] \uparrow$ &85.57 &86.73 &\textbf{89.56} \\
Median vDSC$\left[ \% \right] \uparrow$ &86.54 &86.97 &\textbf{90.33} \\
STD vDSC $\left[ \% \right] \downarrow$ & 6.32 &5.02 &\textbf{3.01} \\
Mean HD $\left[ mm \right] \downarrow$ & 14.44 &11.06 &\textbf{9.20} \\
ABD$\left[ mm \right] \downarrow$ & 1.494 &1.699 &\textbf{1.051} \\
Mean RAVD $\left[ mm \right] \downarrow$ & 19.65 &18.27 &\textbf{11.32} \\
\hline
\end{tabular}
\end{center}

\end{table}
\setlength{\tabcolsep}{1.4pt}

The statistical results of the three methods are shown in Table \ref{table:valdsc}. We use six ways to evaluate the results and all of parameters show our proposed Bridged U-net performs better than stacked U-net and U-net. From the first rows and second row we can see, the average and median vDSC values of our method are highest. Besides, the vDSC standard deviation of Bridged U-net is lowest, demonstrating our method is stable.
Considering Hausdorff distance (HD) , average boundary distance (ABD) and relative absolute volume difference (RAVD) shown in last three rows in Table \ref{table:valdsc}, Bridged U-net suffer lowest value compared with other methods, which means Bridged U-net perform best in these parameters. It can be proved that the proposed method obtains significant improvement on the prostate segmentation compare with stacked U-net and U-net.

\subsection{Testing Results}

\setlength{\tabcolsep}{4pt}
\begin{table}[ht]
\begin{center}
\caption{Quantitative comparison between the proposed method with other methods on testing data.  $\uparrow$ means the higher value is better. $\downarrow$ means the lower value is better.}
\label{table:testdsc}
\begin{tabular}{l|c|c|c|c|c|c}
\hline\noalign{\smallskip}
Team  & DSC$\left[ \% \right]\uparrow$ & HD$\left[ mm \right]\downarrow$ & ABD $\left[ mm \right]\downarrow$ & RAVD $\left[ mm \right]\downarrow$& Score$\uparrow$\\
\noalign{\smallskip}
\hline
\noalign{\smallskip}
Ours(Bridged U-net)&  \textbf{89.96} & 5.5788 &\textbf{1.5938}&7.2674 & \textbf{86.50}\\
DenseFCN(DenseNet)  &  88.98 & \textbf{5.3219} & 1.6619 &\textbf{6.3514} & 86.36 \\
MBIOS(U-net)   &  88.06& 10.1561 & 2.4928 & 7.0986 & 83.66\\
UdeM 2D(ResNet)  &  87.42& 5.8899 & 1.954 & 12.3722& 83.45\\
Ours(Stacked U-net) &  87.15 & 9.6123& 14.5539 & 11.32 & 81.44\\
\hline
\end{tabular}
\end{center}

\end{table}
\setlength{\tabcolsep}{1.4pt}
Our testing results have been submitted to MICCAI PROMISE12 grand-challenge website and evaluated by the organizer. The total score for ranking is obtained after calculating each metrics (mean DSC, HD, ABD, RAVD) by comparing testing result with ground truth segmentation results. The result shows that Bridged U-net performs much better than 2D U-net architecture. We have compared our method with other state-of-the-art 2D methods. Team DenseFCN, using dense block to replace convolution blocks of U-net, was ranked $1^{st}$  in 2D method before we submitted Bridged U-net. Team MIBOS uses U-net architecture while team UdeM 2D uses residual block to replace convolution blocks of U-net. In addition to these methods, we also provide the result of a stacked U-net for reference. From Table \ref{table:testdsc} we can see that our proposed network performs best in DSC and ABD metrics, additionally, we get the highest total score among 2D methods.

\section{Conclusion}
In this paper, we proposed network bridging architecture, which makes stacked U-net suitable for small image datasets such as medical image datasets. We also discussed the bridging and skip connection methods and find out that concatenation is better for network bridging while addition is better for skip connection. Besides of this, we proposed ELU and ReLU combination to improve network performance, which is also effective in traditional image classification task. In addition to activation function, we proposed cos-dice loss to solve the oscillation problem during network training. We use MICAAI PROMISE12 dataset to evaluate our network and the result shows that our network performs better than original U-net, stacked U-net and other state-of-the-art methods.
%






\end{document}